\documentclass[]{article}
\usepackage[letterpaper]{geometry}
\usepackage{amta2016}
\usepackage{times}
\usepackage{url}
\usepackage{latexsym}
\usepackage{natbib}
\usepackage{layout}

\begin{document}

\title{Normalyzing Numeronyms - A NLP approach}  

\author{\name{\bf Avishek Garain} \hfill  \addr{avishekgarain@gmail.com}\\ 
        \name{\bf Sainik Kumar Mahata} \hfill \addr{sainik.mahata@gmail.com}\\ 
        \name{\bf Subhabrata Dutta} \hfill \addr{subha0009@gmail.com}\\ 
        \addr{Computer Science, Jadavpur University}
}

\maketitle
\pagestyle{empty}

\begin{abstract}
  We developed a Statistical Automatic Post-Editing (SAPE) system that works on Machine Translation (MT) output. A hybrid word alignment model was employed into the SAPE system. The proposed hybrid approach combines different word alignment tables and provides the well estimated alignment links to the SAPE system. This also allows the proposed system to correct lexical errors, erroneous words using insertion and deletion, as well as word ordering. We carried out the experiments on parallel dataset consisting of English text, Spanish MT output and corresponding post-edited output.   In this paper, we have also applied the Hierarchical Phrase Based SMT (HPBSMT) to the SAPE system. It has to be mentioned that the output of our SAPE system not only provides better translations than the standard MT output, but also reduces the post-editing efforts as per the evaluation done with respect to different MT evaluation metrics (BLEU, TER and METEOR).
\end{abstract}

%

\section{Introduction}

\label{intro}
In the context of Machine Translation (MT), the quality of the translated text is not always considered good as per human understanding. Post-editing is the process of manually converting the machine translated texts into human readable form. While human post editing process is useful for small sized documents, it is not at all a feasible solution for large sized documents since it involves a lot of human post editing effort and time. 
In general, the term "Post-Editing" (PE) is defined as the correction performed by humans over the translation produced by a MT system~\cite{veale:1997}, often with minimum amount of manual labor~\cite{taus:2010} and as a process of modification rather than revision~\cite{Loffler:1985}.

In order to measure the human post editing effort accurately, we use TER (Translational Error Rate) value~\cite{Snover:2006}. It calculates the number of editing operations, including phrasal shifts that are needed to change a hypothesis translation into an adequate and fluent sentence, normalized by the length of the final sentence. It is observed that the translations provided by current MT systems often fail to deliver desirable output. Thus, to improve the quality, translations are corrected or post-edited by human translators. One of the major goals of an automatic PE system is to reduce the effort of the human post-editors by automatically customizing the MT output to be suited for a particular translation domain. 

In general, MT output suffers from a number of adequacy errors which include incorrect lexical choice, word ordering, insertion, deletion, etc. Thus, in the present attempt, we developed a SAPE system that covers the above-mentioned errors while dealing with the translated and post-edited MT outputs. In addition to SAPE system, we have also incorporated a Hierarchical Phrase-Based SMT (HPBSMT)~\cite{Chiang:2007} which is capable to correct word ordering error. 
The performance of a SMT (as well as SAPE) system heavily relies on bilingual data and word alignment methods. Therefore, we present a hybrid word alignment method in order to provide well estimated word alignment links to be used by our SAPE system. 
On the other hand, during the phrase alignment step, the system automatically estimates the word insertion errors (by considering one-to-many alignment links between MT--PE aligned data), word deletion errors (by considering many-to-one alignment links between MT--PE aligned data). It also handles lexical errors (by estimating lexical weights during model estimation) and word ordering. It has to be mentioned that the HPBSMT facilitates word ordering, as it uses synchronous context free grammar (SCFG)~\cite{Aho:1969} based on hierarchical phrases.    

We evaluated our system by computing the scores using BLEU~\cite{Papineni:2002}, TER~\cite{Snover:2006} and Meteor~\cite{Denkowski:2011} metrics, which show that our SAPE system produces significant improvement over the raw MT output. 
 
The remainder of the paper is organized as follows. 
Section \ref{rw} gives an overview of the related work, Section~\ref{sd} describes the components of our system: preprocessing, hybrid word alignment method and HPBSMT system. 
In Section~\ref{exp}, we present the experimental setup whereas Section~\ref{eval} provides the results with some analysis, followed by conclusions and further work described in Section~\ref{conclude}.

\section{Related Research}
\label{rw}
In the recent trends, various works have been attempted in automated post editing process with the help of Statistical MT (SMT) or Phrase based SMT (PBSMT) systems, as suggested by ~\cite{Lagarda:2009:HLT}. This means that the translation from the source language to the target language can be done using Rule based Machine Translation (RBMT) or PBSMT systems and the output can be fed to another SMT or PBSMT system to get the post edited form. Therefore, one of the advantages of APE systems is that they can adapt any black-box MT engine output and provide automatic PE output without retraining or re-implementing the original MT engine.

APE approaches cover a wide range of methods.
~\cite{Simard:2007:NAACL} and ~\cite{Simard:2007:WMT} applied SMT for post-editing, handled the repetitive nature of errors typically made by rule-based MT systems.  
Similarly, ~\cite{Pal:2015} applied SMT method with enhanced word alignment strategies for automatic post-editing on MT output.
~\cite{Rosa:2012:SMT} and ~\cite{Marecek:2011:WMT} applied a rule-based approach to APE on the morphological level.

~\cite{Knight:1994} argued in favor of using a separate APE module, which is then portable across multiple MT systems and language pairs, and suggested that the post-editing task could be performed using SMT techniques. In connection to that, 
~\cite{Allen:2000} sketched the outline of such an automated post-editing (APE) system, which would automatically learn the post-editing rules from a tri-parallel corpora consisting of source, raw MT and post-edited output. ~\cite{Elming:2006} suggested the use of transformation-based learning in order to automatically acquire rules of error correction from such data; however, the present method is only applied to reduce the lexical choice errors. 

The current post-editing work was done by keeping the concept of sentence structure in mind. By default, English sentences and Spanish sentences follow the SVO (Subject-Verb-Object) structure when composing sentences. Thus, such rule can be considered during post editing. Working with this concept might be easy as it doesn't have the need for SMT or PBSMT systems to be fed with complex rules and makes the process of APE faster. 

~\cite{Denkowski:2015:PhD} developed a method for integrating real time post-edited MT output into a translation model, by extracting a grammar for each input sentence. 
In some cases, the studies have even shown that the quality of MT plus PE can exceed the quality of human translation~\cite{Fiederer:2009,Koehn:2009,DePalma:2009} as well as the productivity~\cite{Zampieri:2014}.
Post-editing can become a MT evaluation method, implying some specific language skills, different from ranking, for which a homogeneous group seems to be enough to perform the task~\cite{Vela:2015}. 

\section{System Description}
\label{sd}
The SAPE system consists of three basic components: pre-processing, a hybrid word alignment model and a HPBSMT based PE system integrated with the hybrid word alignment model. 
The system has been trained on monolingual Spanish MT output as well as its corresponding manually post-edited data collected from the WMT-2015 APE task.   
\subsection{Pre-processing} \label{preprocess}
Initially, we pre-processed the training data in order to enhance the quality of the word alignment model. The Spanish MT output and corresponding manually post-edited versions are tokenized and part of speech  (POS) tags were extracted using TreeTagger\footnote{http://www.cis.uni-muenchen.de/~schmid/tools/TreeTagger/}~\cite{Schmid:1994}. 
We performed several pre-processing steps such as lower casing and POS matching of unigrams before using the word alignment model. We have also performed the POS based pattern alignment model at bigram level. Such aligned POS patterns of bigrams are replaced by their corresponding surface forms only when the alignment is established by METEOR.
The following steps were followed to prepare the bigram level alignment based on POS tags of the machine translated text and the corresponding post edited version.
\begin{itemize}
	\item[1] Initially, the POS tags of each word from every sentence were extracted. This procedure was done for both the machine translated text and the post edited text. Example of the step is given below.
	
	POS MT Text: CommanderX/NC :/COLON Toad/NC su/PPO gilipollas/NC ./FS
	
	POS PE Text: CommanderX/NC :/COLON Sapo/NC eres/VSfin un/ART gilipollas/NC ./FS
	
	\item[2] The POS tags of both the files were converted to bigrams in the next step. The example of the procedure is given below.
	
	Bigram POS MT Text: NC-COLON COLON-NC NC-PPO PPO-NC NC-FS  
	
	Bigram POS PE Text: NC-COLON COLON-NC NC-VSfin VSfin-ART ART-NC NC-FS 
	
	\item[3] The bigrams form both the text files were supplied as input to METEOR alignment. The alignment links in terms of positional index for the above example are given below:
	
	METEOR Alignment: 0-0 1-1 4-5
	
	\item[4] We use the above alignment links to produce MT-PE alignment at bigram level. Finally, we replace the surface forms of the words to produce the surface level alignment of the bigrams. 
	
	$CommanderX : ||| CommanderX :$\\
	$: Toad ||| : Sapo$\\
	$gilipollas . ||| gilipollas$ .	
\end{itemize}

It has been observed that the system is capable to correct wrong lexical choice in MT output(e.g., $: Toad ||| : Sapo$) using the bigram alignment method. Thus, we add these bigram aligned data as addition to the training data for training of the word alignment model.

\subsection{Hybrid Word Alignment} \label{HWA}
The monolingual (Spanish MT output--PE output) word alignment process has been carried out by using the word alignment method based on edit distance (METEOR word aligner~\cite{Lavie:2007}). We also added an additional word alignment method provided by the Berkeley word aligner~\cite{Liang:2006}. 
\subsubsection{METEOR Alignment}
\label{MA}
METEOR is an automatic MT evaluation metric that provides alignment between the hypothesis and reference. 
Given a pair of strings such as $H$ and $R$ to be compared, METEOR initially establishes a word alignment relation between them. 
The alignment is a mapping method between $H$ and $R$, which is built incrementally by the following sequence of word-mapping modules: 
\begin{itemize}
	\item {\bf Exact:} maps if they are exactly same. 
	\item {\bf Porter stemming:} maps if they are same on their stemmed output obtained using the Porter stemmer. 
	\item {\bf WN synonymy:} maps if they are appeared as synonyms in the WordNet.
\end{itemize}
If multiple alignments exist, METEOR selects the alignment for which the word order in two strings is similar (i.e. having the fewest crossing alignment links). The final alignment is produced with respect to H and R by considering the union of all the alignments of different stages (e.g., Exact, Porter Stemming and WN synonymy).

\subsubsection{Berkeley Aligner}
The Berkeley Aligner is used to align words across sentence pairs in a parallel corpus (in these case, the parallel monolingual MT-PE corpus). We initially trained our model on the parallel corpus using the fully unsupervised method of producing Berkeley word alignments. The Berkeley aligner is an extension of the Cross Expectation Maximization word aligner. The aligner uses agreement between two simple sequence-based models by training and facilitating substantial error reductions over standard models. 
Moreover, it is jointly trained with HMM models, and as a result, the AER ~\cite{Vilar:2006} was reduced by 29\%.

\subsubsection{Hybridization} 
The hybrid word alignment method consists of two steps of hybridization:
\begin{itemize}

 \item[1] It combines the statistical word alignment methods like Berkeley word alignment with Grow-Diag-Final-And (GDFA) heuristic ~\cite{Koehn:2010} as well as edit distance based aligner such as METEOR. We rely METEOR word alignment than Berkeley aligner because METEOR provides edit distance based monolingual alignment which is more informative and produce accurate alignment table compare to the Berkeley aligner. 
The additional alignment links which are failed by METEOR are collected from the alignment table provided by the Berkeley aligner.
 
 \item[2] It combines different kind of alignments such as alignment of surface form, parts of speech form, stem form and bigram POS form (as described in ~\ref{preprocess}) provided by the hybridization method 1. We applied union to combine different word alignment tables and hypothesize that all alignments are correct. All the alignment tables are joined together and duplicate entries are removed~\cite{Pal:2013,Pal:2015}.      
\end{itemize}

\subsection{HPBSMT}
The Hierarchical PB-SMT is based on Synchronous Context Free Grammar (SCFG)~\cite{Aho:1969}. SCFG rewrites rules on the right-hand side by aligned pairs ~\cite{Chiang:2007}. 
\begin{equation}\label{eq7}
	X \rightarrow <\gamma, \alpha, \sim>
\end{equation}
where X represents a non-terminal, $\gamma$, $\alpha$ represent sequences of both terminal and non-terminal strings and $\sim$ represents a one-to-one correspondence between the occurrences of non-terminals appearing in $\gamma$ and $\alpha$.   

The weight of each rule is defined as :
\begin{equation}\label{eq8}
	w( X \rightarrow <\gamma, \alpha, \sim>) = \prod_{i} \phi_i(X \rightarrow <\gamma, \alpha, \sim>)^{\lambda_{i}}
\end{equation}

where $\phi_i$ are features defined on each of the rules and $\lambda_{i}$ is the corresponding weight of $\phi_i$. The features are associated with 4 probabilities: frequency
probability $P(\gamma|\alpha)$, $P(\alpha|\gamma)$, lexical frequency probability $P_{w}(\gamma|\alpha)$, $P_{w}(\alpha|\gamma)$ and a Phrase penalty $exp(-1)$.

There exist two additional rules called "glue rule" or "glue grammar" :
\begin{equation}
	S \rightarrow < S X, S X >
\end{equation}
\begin{equation}
	S \rightarrow < X,X >
\end{equation}

These rules are used when no rule could match or the span exceeds a certain length (search depth: set the same as the initial phrase length limit). These rules simply monotonically connect translations of two adjacent blocks together.

The weight of the above type of rule is defined as
\begin{equation}\label{eq8}
	w( S \rightarrow < S X, S X >) = exp(-\lambda_{g})
\end{equation}

where $\lambda_{g}$ controls the model's preference for hierarchical phrases over serial combination of phrases. 

The weight ($w(d_g)$) of the derivation grammar ($d_g$) for generated  source ($f_d$) and target ($e_d$) string is the product of the weights of the rules used in translation $w(r)$, language model probability $P_{lm}$ and the word penalty $exp(-\lambda_{wp}|e|)$ with some control over the length of the target output ($e$). The representation of $d_g$ can be defined as a triplet $<r,i,j>$, where, $r$ stands for grammar rule to rewrite a non-terminal that extends span $f_{d_{i}}^j$ on the source side.
\begin{equation}\label{eq8}
	w(d_g) =\prod_{<r,i,j>\in d_g} w(r) \times P_{lm}^{\lambda_{lm}} \times exp(-\lambda_{wp}|e|)
\end{equation}

\section{Experiments}
\label{exp}
We performed the experiments on the development set and test set provided by the organizers of the APE task in the WMT2015.

\subsection{Data}
The 3-way parallel development set consists of 1,000 sentence triplets containing 21,617 English words in the source, 23,213 words in machine translated Spanish output and 23,098 words in the post-edited translations, respectively.

The texts were provided separately, each English sentence being aligned to the corresponding Spanish MT output and the post-edited MT output.
In case of the training data, the number of sentences was less, having an impact on the number of words.
The number of parallel sentences in the training data was 11,272 having an impact on the number of 238, 335 words in English source text whereas the MT output had 257,644 words and the post-edited MT output had  238,335 words, respectively.
Similarly, the test set provided by the WMT 2015 APE task consists of 1,817 parallel sentences containing a total of 38,244 words on English side and 40,925 words on the corresponding MT output. 
The data sets did not require any pre-processing in terms of encoding or alignment.
The additional monolingual Spanish data obtained from the WMT-2013 archive containing 51,381,432 tokens after filtering. As the monolingual corpus contains some non-Spanish words and sentences, we apply the Language Identifier ~\cite{Shuyo:2010} on monolingual Spanish corpora. 
From the monolingual corpus, we discarded the sentences that belong to different languages or contain different language segments. 

\subsection{Experimental Settings}
The effectiveness of the present work is demonstrated by using HPBSMT model. 
For building our SAPE system, we experimented with respect to various phrase lengths for the translation model and $n$--gram settings for the language model. 
We found that, using a maximum phrase length of 7 for the translation model and a 5-gram language model, the SAPE model produces the best results in terms of BLEU~\cite{Papineni:2002} scores. 

The other experimental settings were concerned with the hybrid word alignment (as described in Section~\ref{sd}) and the phrase-extraction~\cite{Koehn:2003} algorithms. 
The Phrase level pairs that occur only once in the training data are assigned an unduly high probability mass (i.e. 1). To alleviate this shortcoming, we performed smoothing of the phrase table using the Good-Turing smoothing technique~\cite{Foster:2006}.
System tuning was carried out using Minimum Error Rate Training (MERT)~\cite{Och:2003} optimized with k-best MIRA~\cite{Cherry:2012} on a held out development set. 
The 5-gram target language model was trained using KenLM~\cite{Heafield:2011}. 
When the parameters were tuned, the decoding was carried out on the held out test set.

\section{Evaluation}\label{eval}
The evaluation process was carried out using three well known automatic MT evaluation metrics: BLEU, METEOR and TER. We assume that the MT output of the test set provided by the WMT-2015 APE task as our base system translation output and we consider the corresponding PE version as a reference set for the evaluation. We perform the following experiment specified in Table~\ref{tab:1} to systematically improve the baseline system. 
Table~\ref{tab:1} provides a systematic comparison between the baseline system and APE systems with three different evaluation metrics. In all cases, our proposed system performed better. Table~\ref{tab:1} also shows the relative improvement over the baseline system and obtains maximum value with respect to TER.
In Table~\ref{tab:1}, the Basic Moses system with vanilla setting (Giza++ word alignment), the performance has been degraded by 3.46 BLEU point while TER increases 2.0 that signifies poor translation. The similar results were also found with respect to PBSMT system with Berkeley word alignment (i.e., experiment 3). In experiment 4, we apply the PBSMT system with monolingual edit distance based word alignment such as METEOR that also fails to perform better than the baseline system. As PBSMT system fails to achieve our goal, we apply HPBSMT system with Hybrid Word alignment. The system reported in experiment 5, has successfully improved over the base system with respect to all automatic evaluation metrics.  

\begin{table}[h]
	\begin{center}
\begin{tabular}{ c l|ccc }
	\hline
	\textbf{Exp} 	&	 \textbf{Systems}	 	&	\textbf{BLEU} 	&	\textbf{MET}		 &	\textbf{TER} \\ \hline
	1 		& 	Baseline System 	& 	65.90 	& 	74.54	 & 	22.71 \\ 
	2 		& 	Basic Moses 	& 	62.44 	& 	72.80 	 & 	24.71 \\ 
	3 		& 	BA\_PBSMT 		& 	62.52 	& 	72.74	 & 	24.53 \\ 
	4 		& 	MA\_PBSMT 		& 	65.13 	& 	74.13 	 & 	23.11 \\ 
	5 		& 	Hyb\_HPBSMT 	& 	\textbf{66.23} 	& 	\textbf{74.73} 	 & 	\textbf{22.33} \\ \hline
\end{tabular}
\end{center}
\caption{Automatic Evaluation of SAPE system, MET:METEOR.}
\label{tab:1}
\end{table}

\section{Conclusion}
\label{conclude}
Our English–-Spanish APE system has successfully improved the translation quality in terms of automatic evaluation metrics. The hybrid word alignment plays crucial role in this tasks. Edit-distance based monolingual aligner provides very well alignment link for our SAPE system. We achieve considerable improvement over the base system after incorporating the hybrid word alignment into the state-of-the-art HPBSMT pipeline.

\bibliographystyle{apalike}
\bibliography{amta2016}

\end{document}